\documentclass[10pt,twocolumn,letterpaper]{article}

\usepackage{cvpr}              
\definecolor{cvprblue}{rgb}{0.21,0.49,0.74}
\usepackage[pagebackref,breaklinks,colorlinks,allcolors=cvprblue]{hyperref}
\usepackage{xcolor}       
\usepackage{colortbl}     
\usepackage{booktabs}     
\usepackage{pgf}          
\usepackage{graphicx}     
\usepackage{multirow}     
\definecolor{heatred}{HTML}{C1272D}
\definecolor{heatblue}{HTML}{0072BC}

\newcommand{\databarRatio}[2][gray!30]{%
  \pgfmathsetmacro{\barwidth}{#2/1.0*1.8}%
  {\color{#1}\rule{\barwidth cm}{1.6ex}}%
  \hspace*{-\barwidth cm}%
  \makebox[1.8cm][l]{#2}%
}

\newcommand{\databarPSNR}[2][gray!30]{%
  \pgfmathsetmacro{\barwidth}{#2/45.0*1.8}%
  {\color{#1}\rule{\barwidth cm}{1.6ex}}%
  \hspace*{-\barwidth cm}%
  \makebox[1.8cm][l]{#2}%
}

\def\paperID{7020} 
\def\confName{CVPR}
\def\confYear{2026}

\title{RecoverMark: Robust Watermarking for Localization and Recovery of Manipulated Faces}

\author{
Haonan An$^1$ \quad Xiaohui Ye$^{3}$ \quad Guang Hua$^{2}$\footnotemark[1] \quad Yihang Tao$^1$ \quad
Hangcheng Cao$^1$ \\  Xiangyu Yu$^3$ \quad Yuguang Fang$^1$\footnotemark[1]\\
$^1$City University of Hong Kong, Hong Kong \quad
$^2$Singapore Institute of Technology, Singapore 828608 \\
$^3$South China University of Technology \\
{\tt\small 
(haonanan2-c, yihang.tommy)@my.cityu.edu.hk, }
{\tt\small 
(hangccao, my.fang)@cityu.edu.hk,
} \\
{\tt\small 
ghua@ieee.org,
}
{\tt\small 
eeyxh2023@mail.scut.edu.cn,
}
{\tt\small 
yuxy@scut.edu.cn
}
}

\begin{document}

\twocolumn[{
\maketitle
\begin{center}
    \vspace{-6mm}
    \captionsetup{type=figure}
    \includegraphics[width=1\textwidth]{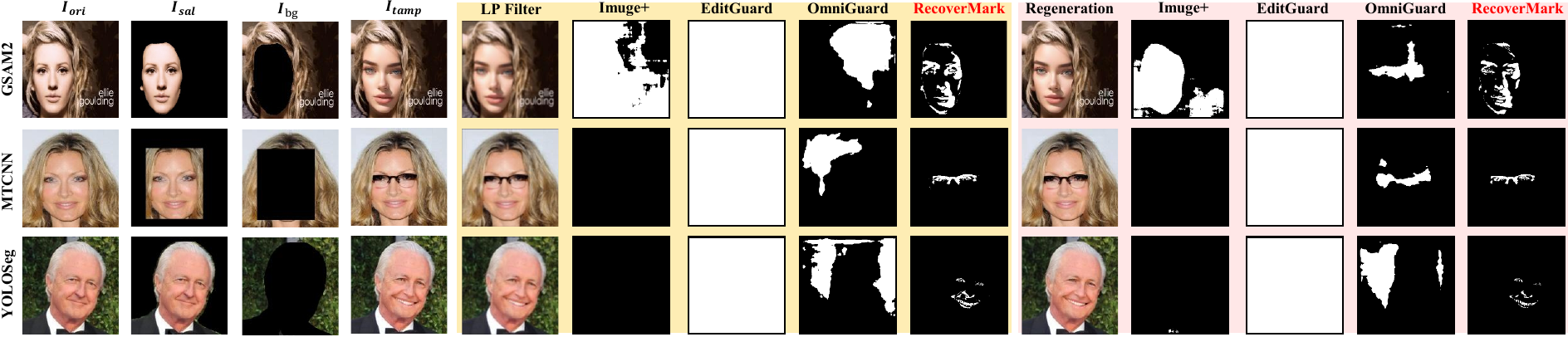}
    \vspace{-7mm}
    \captionof{figure}{Visualization of face manipulation localization for Imuge+~\cite{ProactiveImmunePlus}, EditGuard~\cite{BaselineEditGuard}, OmniGuard~\cite{BaselineOmniGuard}, and our proposed RecoverMark. The evaluation is conducted using three distinct region-of-interest definitions: GSAM2~\cite{gsam} (face-only segmentation), MTCNN~\cite{mtcnn} (head bounding box), and YOLOSeg~\cite{yoloseg} (full-body). All methods are tested against traditional low-pass filtering and an advanced regeneration attack~\cite{attack_regeneration}, where $I_{\text{ori}}$ denotes the original image, $I_{\text{sal}}$ represents segmented saliency (face), $I_{\text{bg}}$ refers to segmented background, and $I_{\text{tamp}}$ stands for the tampered image.} 
    \label{fig:visualization_compare}
\end{center}}
]

\renewcommand{\thefootnote}{\fnsymbol{footnote}}
\footnotetext[1]{Corresponding authors.}
\begin{abstract}
The proliferation of AI-generated content (AIGC) has facilitated sophisticated face manipulation, severely undermining visual integrity and posing unprecedented challenges to intellectual property (IP). In response, a common proactive defense leverages fragile watermarks to detect, localize, or even recover manipulated regions. However, these methods always assume an adversary unaware of the embedded watermark, overlooking their inherent vulnerability to watermark removal attacks. Furthermore, this fragility is exacerbated in the commonly used dual-watermark strategy that adds a robust watermark for image ownership verification, where mutual interference and limited embedding capacity reduce the fragile watermark's effectiveness.
To address the gap, we propose RecoverMark, a watermarking framework that achieves robust manipulation localization, content recovery, and ownership verification simultaneously. Our key insight is twofold. First, we exploit a critical real-world constraint: an adversary must preserve the background's semantic consistency to avoid visual detection, even if they apply global, imperceptible watermark removal attacks. Second, using the image's own content (face, in this paper) as the watermark enhances extraction robustness. Based on these insights, RecoverMark treats the protected face content itself as the watermark and embeds it into the surrounding background. By designing a robust two-stage training paradigm with carefully crafted distortion layers that simulate comprehensive potential attacks and a progressive training strategy, RecoverMark achieves a robust watermark embedding in no fragile manner for image manipulation localization, recovery, and image IP protection simultaneously. Extensive experiments demonstrate the proposed RecoverMark's robustness against both seen and unseen attacks and its generalizability to in-distribution (ID) and out-of-distribution (OOD) data. Code will be released upon acceptance.

\end{abstract}

\section{Introduction}
\label{sec:intro}
Recent advances in AI-generated content (AIGC) models, such as numerous variants of Stable Diffusion \cite{StableDiffusion} and GANs \cite{GAN, StyleGAN}, have empowered unprecedented creativity and enabled the realization of human imagination. However, these powerful techniques are double-edged swords, as they can also be misused for malicious image manipulation. In particular, facial content editing has become one of the most notorious applications, raising serious concerns in morality \cite{ConcernMorality}, finance \cite{ConcernFinance}, and copyright protection \cite{ConcernCopyright}. Such threats not only pose a formidable challenge to media authentication but also affect the integrity of broader intelligent systems \cite{11127818, tao2025gcpguardedcollaborativeperception, hucpguard2025, li2026splitcom}.

Previous countermeasures have primarily focused on passively detecting and localizing anomalies introduced by image manipulations \cite{Passive_salloum2018image, PassiveMantraNet, PassiveDraw, Passive_kong2022detect}, such as visual artifacts \cite{Passive_VisualArtifacts}, local noise inconsistencies \cite{Passive_NoiseInconsis}, and resolution discrepancies \cite{Passive_Reso_disc}. However, these methods often depend on prior knowledge of potential manipulation techniques, which limits their generalization to unseen or novel types of forgeries \cite{PassiveMVSSNet, BaselineEditGuard}. To address this issue, proactive frameworks for manipulation detection and localization have been developed \cite{ProactiveImmune, Proactive_asnani2022proactive,ProactiveImmunePlus}. State-of-the-art methods \cite{BaselineEditGuard, BaselineOmniGuard} in this area employ dual-watermark frameworks that combine both fragile and robust watermarks. A fragile watermark is embedded into the image before distribution and, after any manipulation, serves as a sensitive indicator to locate tampered regions. In contrast, a robust watermark, which is resistant to common image processing operations, is used concurrently for applications such as copyright authentication.

However, these methods suffer from several critical weaknesses. \textbf{1)} They rely on an implicit assumption that the image manipulator is unaware of the presence of a fragile watermark. This assumption is rather strong in practical scenarios, where an adversary can intentionally perform image post-processing operations for watermark removal attacks before or after carrying out malicious manipulations. As illustrated in Figure~\ref{fig:visualization_compare}, such removal attacks (both traditional low-pass filter and advanced regeneration attack~\cite{attack_regeneration}) can readily corrupt the embedded fragile watermark, thereby rendering the manipulation detection and localization ineffective. \textbf{2)} In dual-watermark frameworks, the co-embedding of a robust watermark alongside the fragile one further weakens the fragile watermark's effectiveness against removal attacks, due to mutual interference between the watermarks as well as the inherent constraints on watermark capacity. \textbf{3)} Existing methods usually neglect an equally critical aspect of watermark: the ability to recover the original content within tampered regions, which is vital for maintaining content integrity \cite{ProactiveImmune, ProactiveImmunePlus}.

\begin{figure}[!t]
    \centering
    \includegraphics[width=0.95\linewidth]{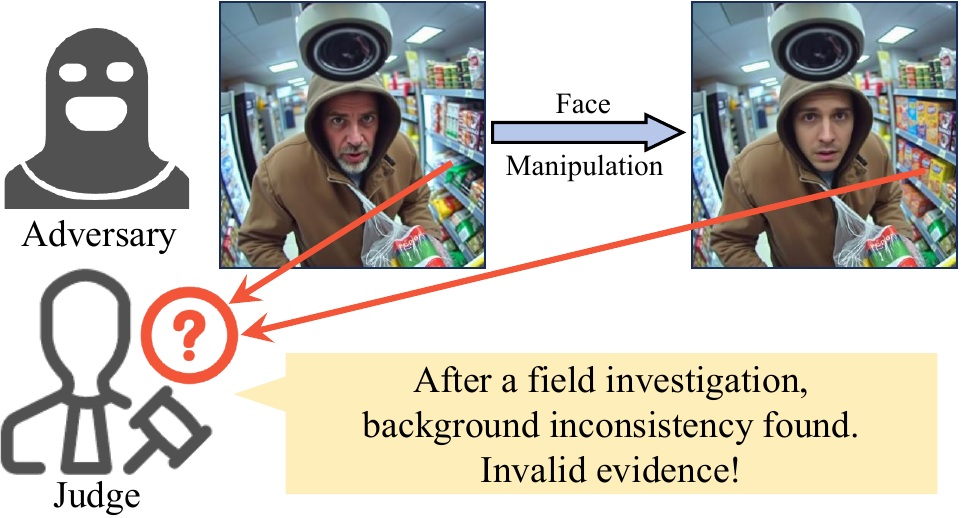}
    \caption{A practical example when a judge determines whether the evidence is credible based on background consistency.}
    \label{fig:assumption}
\end{figure}

To address these limitations, we propose RecoverMark, a robust watermarking framework that integrates manipulated region localization, original content recovery, and ownership verification, with a particular focus on face tampering. Our key insight is that, in practical scenarios,
adversaries must preserve the background's semantic consistency, which implies any modification should be imperceptible to human perception (this constraint does not prevent them from applying imperceptible post-processing to the background), to avoid being visually detected through trivial comparisons with the surrounding environment. For instance, as illustrated in Figure~\ref{fig:assumption}, consider an attacker attempting to evade allegations by manipulating facial regions in surveillance footage. While the face can be altered, significant background modification is infeasible, as this would introduce noticeable inconsistencies with the real environment. Such artifacts would be readily detected during judicial review, rendering the manipulated evidence unreliable. Leveraging this observation, RecoverMark segments facial content and considers it as a watermark. A two-stage, progressively trained embedding pipeline then embeds this facial watermark into the background in a robust manner, ensuring resilience against common image post-processing operations (e.g., JPEG compression, Gaussian noise, patch removal) as well as advanced removal attacks, such as the lattice attack \cite{Liu2023Erase_Removal_Attack} and the regeneration attack \cite{attack_regeneration}. After adversarial face manipulations, RecoverMark can faithfully extract the original facial information, enabling both precise tamper localization and accurate content recovery. Moreover, since the facial watermark is embedded robustly, it also serves as a copyright verification mechanism, mitigating the vulnerabilities inherent in prior dual-watermark frameworks. Our main contributions are as follows: 
\begin{itemize}
    \item We reveal the weaknesses of the state-of-the-art proactive image manipulation localization methods and propose RecoverMark, a robust watermarking framework that simultaneously supports tamper localization, content recovery, and copyright verification, with a particular focus on face manipulation scenarios.
    \item We introduce a two-stage progressive training pipeline that treats facial content as a watermark and embeds it into the background in a robust manner, ensuring resilience against both common post-processing and advanced watermark removal attacks.
    \item We conduct extensive qualitative and quantitative experiments by using in-distribution (ID) and out-of-distribution (OOD) datasets under seen and unseen removal attacks, demonstrating the generalizability, robustness, and effectiveness of RecoverMark.
\end{itemize}

\section{Related Work}
\subsection{Passive Image Manipulation Localization}
Passive methods detect inconsistencies between authentic and tampered regions using DNNs \cite{Passivezhou2018learning, Passiveli2019localization,PassiveMantraNet, PassiveDOAGAN}. For instance, CFL-Net \cite{PassiveCFL-Net} uses contrastive learning to maximize feature discrepancy, while MVSS-Net \cite{PassiveMVSSNet} introduces multi-view and multi-scale learning to capture robust, semantic-agnostic traces. Building upon these, MPC \cite{PassiveMPC} integrates the advantages of both methods by proposing a Multi-view Pixel-wise Contrastive (MPC) algorithm. Additionally, HiFi-Net \cite{PassiveHiFiNet} and DiffForensics \cite{PassiveDiffforensics} enhance generalization by learning manipulation taxonomies and leveraging diffusion model priors, respectively. Despite these advancements, passive methods exhibit inherent limitations. Their performance degrades significantly when forensic inconsistencies in tampered regions are subtle. Furthermore, their effectiveness is often diminished in real-world applications, as crucial forensic traces can be obscured or erased by unpredictable post-processing operations during media transmission.

\subsection{Proactive Image Manipulation Localization}
Proactive methods share a common principle: the performance of image manipulation localization can be enhanced by embedding imperceptible redundancy into an image without significantly degrading its visual quality \cite{ProactiveRuiz2020disrupting, ProactiveYeh2020disrupting, ProacitveWang2021faketagger, ProactiveRWN, ProactiveZhao2024proactive}. For instance, Asnani et al. \cite{Proactive_asnani2022proactive} introduces a method that embeds a learned, imperceptible template into images to significantly improve the detection of manipulations and generalization against unseen AI models. Similarly, Imuge and its advanced version Imuge+ \cite{ProactiveImmune, ProactiveImmunePlus} embed invisible noise into an image to precisely localize tampered regions. Furthermore, EditGuard \cite{BaselineEditGuard} and its updated version OmniGuard \cite{BaselineOmniGuard} discovers that fragile watermarks, upon extraction, leave tampering artifacts at the corresponding manipulated locations. This enables precise localization of tampered areas and has achieved state-of-the-art performance in the field. However, these proactive methods are fundamentally built upon fragile watermarking, making them inherently vulnerable to watermark removal attacks, where an adversary can attempt to remove the embedded watermark before or after performing the actual manipulation.

\subsection{Self-Recovery after Manipulation}
Prior to the deep learning era, some methods for image manipulation address both localization and self-recovery \cite{RecoveryZhang2009fragile,RecoveryZhang2010reference, RecoveryZhang2011watermarking}, typically by self-embedding transformed data like DCT coefficients \cite{RecoveryHe2006wavelet, RecoveryZhang2008fragile}. However, these techniques are often limited by insufficient embedding capacity. While the focus in the DNN era has shifted primarily to manipulation localization, a few recent works have revisited self-recovery. For instance, Imuge/Imuge+ \cite{ProactiveImmune, ProactiveImmunePlus} introduces the first DNN-based framework for joint localization and recovery, and CoAtNet \cite{RecoveryPalani2024semi} proposes a two-level embedding mechanism only for medical images. A fundamental limitation shared by both traditional and recent self-recovery approaches is their reliance on fragile self-embedding, which makes them highly vulnerable to performance degradation from common post-processing attacks.

\section{Preliminary}

\subsection{Tampering Localization by Fragile Watermark}
To better understand the motivation behind our proposed RecoverMark, we first review the high-level workflow of prior fragile-watermark-based tampering region localization methods \cite{BaselineEditGuard, BaselineOmniGuard} and analyze their limitations. Considering an image or a video frame $I_{\text{ori}}$ under the threat of face manipulation, fragile-watermark-based methods first embed the fragile watermark into $I_{\text{ori}}$:
\begin{equation}
    I_{\text{cntr}} = HNet(I_{\text{ori}} , I_{\text{fw}}),
\end{equation}
where $I_{\text{cntr}}$ is the watermarked (container) image, $HNet$ denotes the watermark hiding network, and $I_{\text{fw}}$ is the fragile watermark image of the same size as $I_{\text{ori}}$. After face manipulation, the tampered $I_{\text{cntr}}$ can be formulated as:
\begin{equation}
    I_{\text{cntr}}' = \mathcal{N}\left(I_{\text{cntr}} \odot \left(I-M\right) + \mathcal{FM} \left(I_{\text{cntr}}\odot M\right)\right),
\end{equation}
where $I_{\text{cntr}}'$ is the manipulated container, $\odot$ refers to Hadamard product, $\mathcal{N}$ stands for the potential noise, $M$ is the binary mask of the face region, and $\mathcal{FM}$ represents the face manipulation operator. The embedded fragile watermark is then extracted by extraction network $ENet$:
\begin{equation}
    I_{\text{fw}}' = ENet(I_{\text{cntr}}').
\end{equation}
Finally, the tampering localization mask $M_{\text{loc}}$ is derived by comparing the extracted watermark with the original one:
\begin{equation}
    M_{\text{loc}} = \mathbf{1}(|I_{\text{fw}}'-I_{\text{fw}}| > \text{threshold}),
\end{equation}
where $\mathbf{1}$ is the indicator function. However, the types of $\mathcal{N}$ considered in these methods are typically limited to traditional image post-processing operations, such as JPEG compression and Gaussian blur, while more advanced and well-studied watermark removal attacks are often overlooked. Moreover, the fragility of the embedded watermark makes it inadequate for reliable ownership verification, necessitating an additional non-fragile copyright watermark. This dual-watermark approach, however, further compromises the effectiveness of fragile watermarking due to the inherent limited capacity for multiple watermark embeddings. Additionally, the embedded $I_{\text{fw}}$ is independent of $I_{\text{ori}}$, rendering it incapable of recovering the original content after tampering.

\subsection{Threat Model}
We consider two adversarial parties in our threat model:

\textbf{Scrutinizer.} The scrutinizer has access to both the original image and the potentially manipulated version produced by the face manipulator. Before distribution, the scrutinizer proactively embeds watermarks into the original image, taking into account the risk of potential watermark removal attacks. Given a suspicious image, the scrutinizer first determines whether the facial region has been tampered with. If tampering is detected, the scrutinizer further localizes the manipulated areas of the face and restores the original content in the tampered regions.

\textbf{Face Manipulator.} The face manipulator has access to the watermarked image produced by the scrutinizer, seeking to alter the facial region (e.g., face swap, attribute change, and expression change) while evading detection. To do this, they may apply watermark removal attacks, ranging from common image processing to deep learning reconstruction, either before or after the manipulation.


\begin{figure*}[!t]
    \centering
    \includegraphics[width=1.0\linewidth]{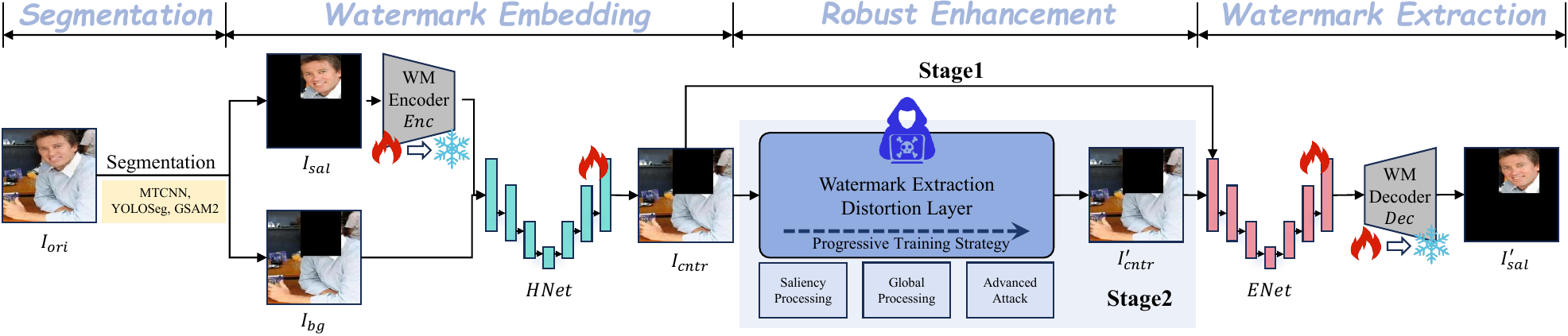}
    \caption{Flowchart of the training process for the proposed RecoverMark, where MTCNN \cite{mtcnn} is used as segmentation model, and the watermark encoder/decoder are trained in Stage $1$ but remain frozen during Stage $2$.}
    \label{fig:workflow}
\end{figure*}

\section{Proposed Method: RecoverMark}
\label{sec:methodology}
In this section, we first present the overall workflow of RecoverMark. Next, we detail the two-stage training pipeline along with progressive training strategy. Finally, we describe how RecoverMark is applied to localize and recover tampered regions while simultaneously satisfying copyright verification requirements.

\subsection{Overview}
\label{sec:overview}
The components and the training workflow of the proposed RecoverMark are illustrated in Figure~\ref{fig:workflow}. The process begins by segmenting the original image $I_{\text{ori}}$ into a saliency $I_{\text{sal}}$ (which, in our case, corresponds to the human face content) and a background region $I_{\text{bg}}$, using  advanced segmentation tools such as MTCNN~\cite{mtcnn}, YoloSeg~\cite{yoloseg}, and SAM2~\cite{sam2}. Prior to watermark embedding via the hiding network $HNet$, a watermark encoder $Enc$ compresses $I_{\text{sal}}$ into a latent representation to better accommodate the limited watermark capacity of $I_{\text{bg}}$. The output of $HNet$ is the watermarked background, referred to as the container image $I_{\text{cntr}}$. In the first training stage, $I_{\text{cntr}}$ is fed directly into the extraction network $ENet$ to extract the latent representation of the embedded watermark, which is further processed by watermark decoder $Dec$ to obtain the extracted watermark $I_{\text{sal}}'$. In the second stage, with both the watermark encoder and decoder frozen, $I_{\text{cntr}}$ undergoes distortion through a dedicated distortion layer, resulting in a perturbed version $I_{\text{cntr}}'$, which is then input to $ENet$. The distortion layer incorporates three primary types of perturbations: saliency processing (which applies noise specifically to the salient region to discourage saliency-dependent extraction), global processing (which perturbs the entire image), and advanced attacks (such as regeneration attacks that simulate sophisticated watermark removal attacks). Together, these distortions encompass nearly all types of distortions likely to occur in real-world settings.
Moreover, akin to human learning, which requires gradual acquisition of knowledge, RecoverMark also benefits from progressive exposure to distortions rather than learning them all at once. Accordingly, we introduce a progressive training strategy during the second stage, which sequentially incorporates these distortions into the training process. Further implementation details are provided in the subsequent sections.
\subsection{Two-Stage Training Pipeline}
\subsubsection{Stage 1: Initial Training}
We first segment an image or a video frame $I_{\text{ori}}$ as:
\begin{equation}
    \label{eq:segmentation}
    I_{\text{ori}} = I_{\text{sal}} + I_{\text{bg}},
\end{equation}
where $I_{\text{sal}}$ represents the salient region of interest (human face in our case), and $I_{\text{bg}}$ corresponds to the remaining background content. Then, we formally formulate the watermark embedding process as:
\begin{equation}
    I_{\text{cntr}} = HNet(\text{Concat}(Enc(I_{\text{sal}}),I_{\text{bg}})),
\end{equation}
where $\text{Concat}(\cdot)$ represents channel-wise concatenation. The corresponding watermark extraction process is defined as:
\begin{equation}
    \label{eq:wm_extration}
    I_{\text{sal}}' = Dec(ENet(I_{\text{cntr}})).
\end{equation}
During the first training stage, all networks, including the watermark encoder $Enc$, the hiding network $HNet$, the extraction network $ENet$, and the watermark decoder $Dec$, are set to training mode, supervised by the following three loss functions. The first one is the fidelity loss:
\begin{equation}
    \label{eq:fidelity_loss}
    \mathcal{L}_{\text{Fidelity}}  = \sum\|I_{\text{cntr}} - I_{\text{bg}}  \|_2^2,
\end{equation}
which constrains the visual distortion introduced during watermark embedding, ensuring that the container image $I_{\text{cntr}}$ maintains high perceptual fidelity to the original background $I_{\text{bg}}$. The second one is the watermark loss:
\begin{equation}
    \label{eq:wm_loss}
    \mathcal{L}_{\text{wm}} = \sum\|I_{\text{sal}}' - I_{\text{sal}} \|_2^2,
\end{equation}
which enforces the accurate extraction and decoding of the embedded saliency content. To further enhance the system's robustness and prevent false watermark detection, we introduce a clean loss defined as:
\begin{equation}
    \label{eq:clean_loss}
    \mathcal{L}_{\text{clean}} = \sum\|Dec(ENet(I_{\text{bg}})) - I_{\text{white}} \|_2^2,
\end{equation}
where $I_{\text{white}}$ is an all-white image, indicating no watermark is embedded. We jointly train these four networks by minimizing the following objective function:
\begin{equation}
    \label{eq:sum_loss}
    \mathcal{L}_{\text{sum}} = \alpha_1 \mathcal{L}_{\text{fidelity}} + \alpha_2 \mathcal{L}_{\text{wm}} + \alpha_3 \mathcal{L}_{\text{clean}},
\end{equation}
where $\alpha_1, \alpha_2, \alpha_3$ are weighting parameters.

\subsubsection{Stage 2: Robustness Enhancement Training}
Stage 2 introduces a distortion layer between $HNet$ and $ENet$ to simulate adversary's behavior and strengthen RecoverMark’s robustness against watermark removal. As $Enc$ and $Dec$ have already acquired stable compression and reconstruction abilities in Stage 1, they remain frozen in this phase. The watermark extraction (\ref{eq:wm_extration}) is thus updated to operate on the distorted container image $I_{\text{cntr}}'$:
\begin{equation}
    I_{\text{sal}}' = Dec(ENet(I_{\text{cntr}}')).
\end{equation}
Despite the altered input caused by distortions, the training objectives remain minimizing (\ref{eq:sum_loss}), since the distortion layer is considered an external perturbation rather than a change in the learning targets.

The design of the distortion layer is critical, as a comprehensive simulation of diverse attack types significantly enhances the general robustness of RecoverMark. However, overly complex distortions can destabilize the training process and hinder convergence. Therefore, it is essential to first classify these distortions to understand their individual properties. Drawing upon existing literature, we categorize the simulated distortions into three groups:
\begin{itemize}
    \item Saliency processing: We add random Gaussian noise to the salient region of $I_{\text{cntr}}$, so that face  extraction does not rely on specific facial information but instead only depends on background.
    \item Global processing: Common global image post-processing operations typically do not introduce noticeable inconsistencies in the background. To simulate such distortions, we apply JPEG compression and global Gaussian noise. Additionally, we observe that, due to the fidelity loss, the watermark’s latent representation is often embedded in high-frequency regions, such as edges of the original image \cite{Boxfree_FrequencyMatters, Boxfree_Zhang_2024}. To account for this, we also incorporate a low-pass filter during distortion.
    \item Advanced attack: We include an state-of-the-art regeneration attack \cite{attack_regeneration}, which can achieve nearly $100\%$ watermark removal success in their experiments. By training to withstand this strongest form of attack, the model is expected to generalize and maintain robustness against other less challenging distortions.
\end{itemize}
To enhance training stability, instead of applying all distortions from the beginning, we employ a progressive training strategy. Training starts with the most challenging regeneration attack, which occupies half of the total epochs. The remaining distortions are then introduced sequentially, each allocated an equal portion of the remaining half of the training epochs. As we will discuss in Section \ref{sec:ablation_study}, we note that the ordering is crucial in our experiments: introducing the regeneration attack first substantially enhances robustness, whereas delaying it prevents RecoverMark from effectively defending against the strongest attack.

\begin{figure}[!t]
    \centering
    \includegraphics[width=0.95\linewidth]{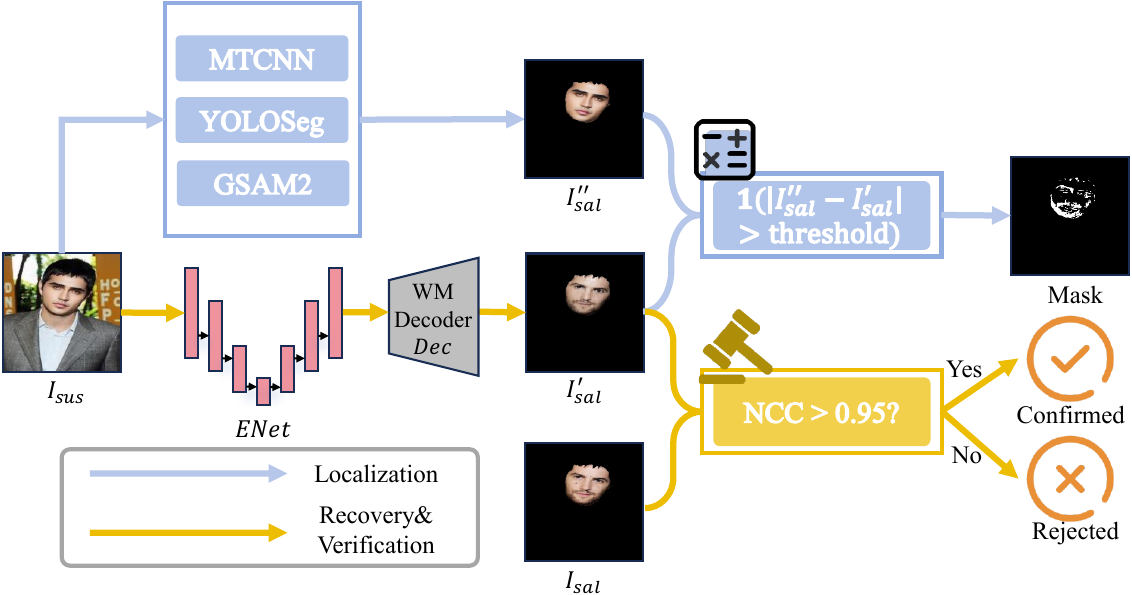}
    \caption{Demonstration of manipulation localization, recovery, and ownership verification for a suspicious image.}
    \label{fig:verification}
\end{figure}

\subsection{Recovery, Localization, and Copyright Verification}
\label{sec:recovery_localization_verification}
As illustrated in Figure~\ref{fig:verification}, after recovering the hidden face information $I_{\text{sal}}'$ from a suspicious image $I_{\text{sus}}$, RecoverMark proceeds with two distinct stages: manipulation localization and copyright verification. For localization, we identify tampered regions by comparing $I_{\text{sal}}'$ with the saliency of $I_{\text{sus}}$ and consequently generate a difference mask. For copyright verification, $I_{\text{sal}}'$ is compared with the original face $I_{\text{sal}}$. Ownership is confirmed if the Normalized Correlation Coefficient (NCC), a metric widely used in watermarking literature \cite{Boxfree_Zhang_2024, Boxfree_DGS}, exceeds $0.95$.

\section{Experimental Results}
\subsection{Experimental Settings}
\subsubsection{Datasets}
We conduct our experiments on two public human face datasets. We use CelebA \cite{Dataset:CelebA} for both training and in-distribution (ID) testing. To evaluate the generalization capability of RecoverMark, we employ FFHQ \cite{Dataset:FFHQ} as an out-of-distribution (OOD) test set. All images are set to $256 \times 256$ in RGB format.

\subsubsection{Evaluation Metrics} 
We assess our method's performance across three tasks. For manipulation localization, we measure the F1 score and the Area Under the Curve (AUC). For facial information recovery, we use the Peak Signal-to-Noise Ratio (PSNR) and the Multiscale Structural Similarity Index (MS-SSIM) \cite{Metric:MS-SSIM}. Finally, for ownership verification, we define the Success Rate as the percentage of images where the NCC between $I_{\text{sal}}'$ and $I_{\text{sal}}$ exceeds the $0.95$ threshold.

\subsubsection{Implementation Details}
For our network architectures, we employ a UNet \cite{arc:UNet} for $HNet$ and use CEILNet \cite{arc:CEILNet} for the $Enc$, $Dec$, and $ENet$. Both training stages are optimized using the Adam optimizer with a learning rate of $0.0002$. The loss weighting parameters $\alpha_{1}$, $\alpha_{1}$, and $\alpha_{3}$ in (\ref{eq:sum_loss}) are all set to $1$ and the threshold for binary localization mask generation is set to $0.15$. Additionally, all experiments are conducted on a server with an NVIDIA GTX $5880$ GPU.

\begin{figure}[!t]
    \centering
    \includegraphics[width=0.95\linewidth]{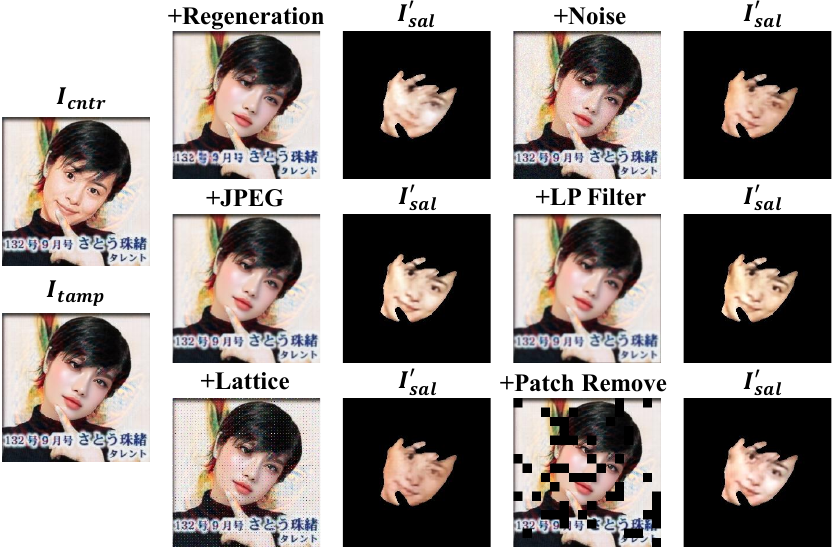}
    \caption{Demonstration of RecoverMark's robustness against a range of attacks (from single to multiple), including the unseen lattice attack \cite{Liu2023Erase_Removal_Attack}, arranged from top to bottom and left to right..}
    \label{fig:recovermark_vis}
\end{figure}

\subsection{Qualitative Evaluation}
We first present the qualitative evaluation of RecoverMark. As illustrated in Figure~\ref{fig:visualization_compare}, before face manipulation, watermark removal attacks such as low-pass filtering and regeneration attack are employed and do not significantly modify the background. While all proactive baselines, including Imuge+ \cite{ProactiveImmunePlus}, EditGuard \cite{BaselineEditGuard}, and OmniGuard \cite{BaselineOmniGuard}, fails to localize the manipulation, our proposed RecoverMark succeeds in precise pixel-level localization. We demonstrate RecoverMark's robustness to a sequence of cumulative attacks in Figure~\ref{fig:recovermark_vis}. The attack sequence commences with a regeneration attack, followed by noise, JPEG compression, a low-pass filter, the lattice attack \cite{Liu2023Erase_Removal_Attack}, and finally a patch removal attack. It is particularly noteworthy that RecoverMark maintains its robustness even against the lattice attack, which was not included in the distortion layer during training, highlighting its strong generalization capabilities.

\subsection{Quantitative Evaluation}
\label{sec:quantitative_evaluation}
\textbf{Robustness and Generalizability.} As shown in \mbox{Tables \ref{tab:CelebA_Structpix2pix} -- \ref{tab:FFHQ_sd_inpaint}}, we benchmark RecoverMark against passive (MVSS-Net \cite{PassiveMVSSNet}, HiFiNet \cite{PassiveHiFiNet}) and proactive (Imuge+ \cite{ProactiveImmunePlus}, EditGuard \cite{BaselineEditGuard}, OmniGuard \cite{BaselineOmniGuard}) methods on ID and OOD datasets. The evaluation covers manipulations from Structpix2pix \cite{brooks2022instructpix2pix} and Stable Diffusion \cite{StableDiffusion} inpaint under a wide range of attacks, including the unseen lattice attack. The results show that RecoverMark maintains exceptional robustness and generalization, significantly outperforming all baselines in localization and recovery metrics across most test conditions.

\begin{figure}[t]
    \centering
    
    \begin{subfigure}[b]{0.48\columnwidth}
        \centering
        \includegraphics[width=\linewidth]{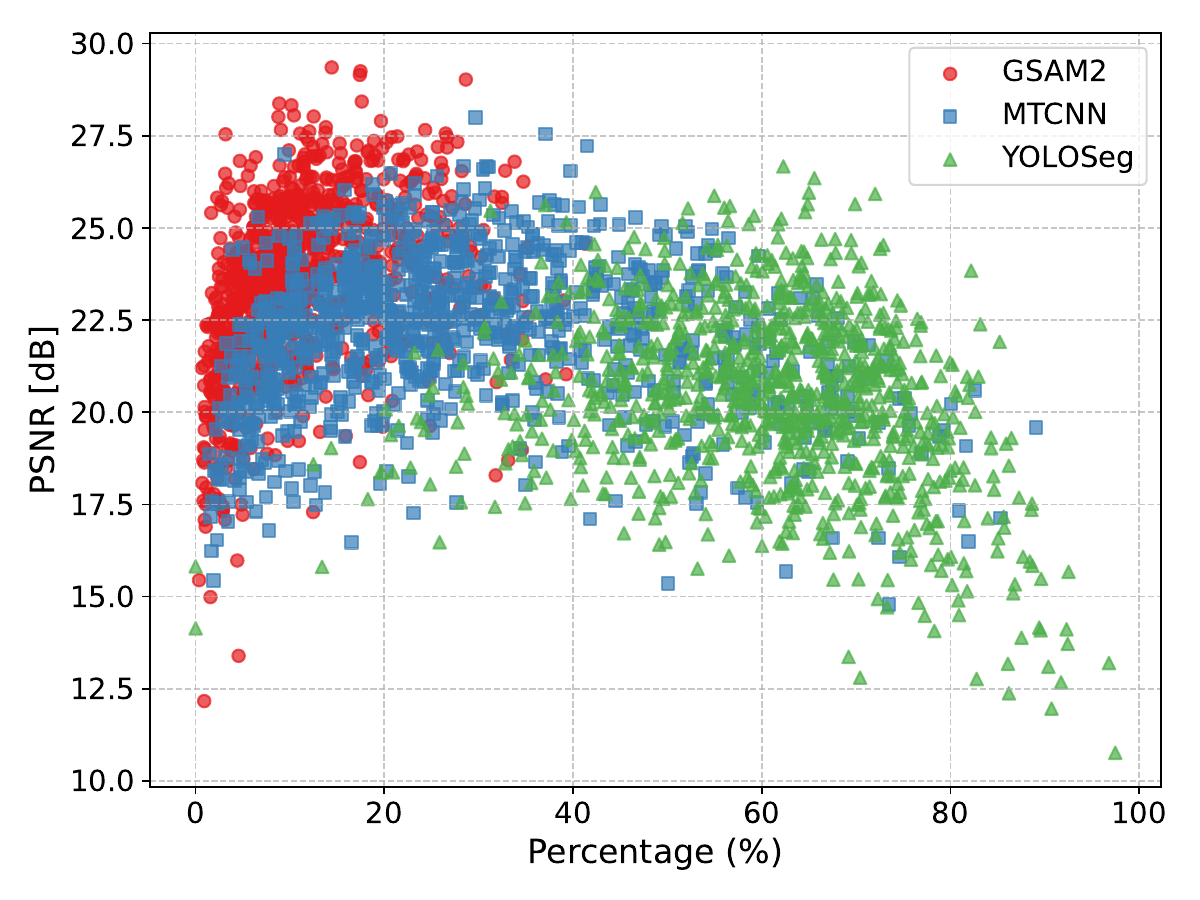}
        \caption{Saliency Fidelity}
        \label{fig:capacity_sub1}
    \end{subfigure}%
    \hfill 
    \begin{subfigure}[b]{0.48\columnwidth}
        \centering
        \includegraphics[width=\linewidth]{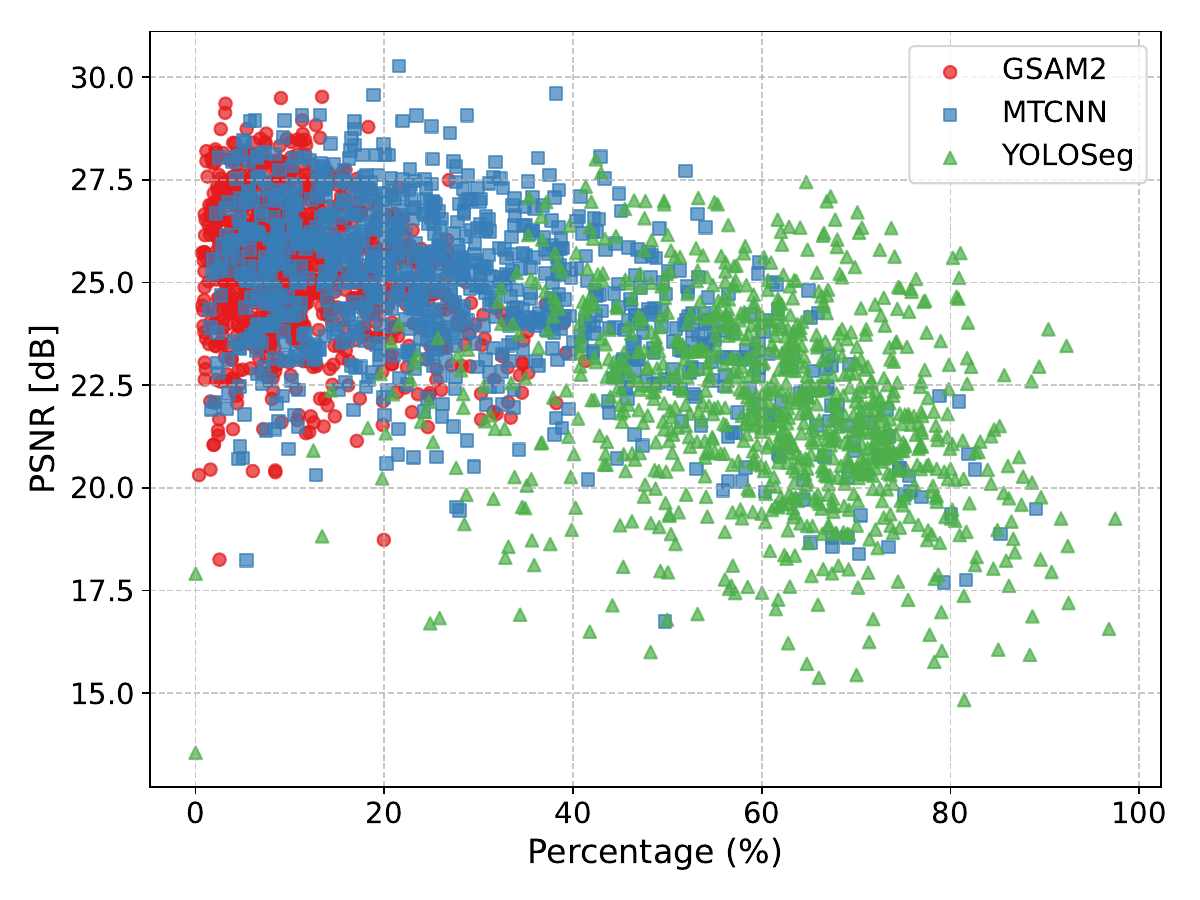}
        \caption{Background Fidelity}
        \label{fig:capacity_sub2}
    \end{subfigure}
    
    \caption{Empirical watermark capacity evaluated by the fidelity of recovered saliency and watermarked background at varying saliency-to-image percentages, measured under regeneration attack.}
    \label{fig:capacity}
\end{figure}

\noindent \textbf{Capacity Evaluation.} We now evaluate the watermark capacity of our proposed method. As illustrated in Figure~\ref{fig:capacity}, subfigures (a) and (b) show the fidelity of the recovered saliency $I_{\text{sal}}'$ and the watermarked background $I_{\text{cntr}}$, respectively, under the regeneration attack as the saliency-to-image percentage varies. The results show a clear trend that as the facial region percentage increases, the fidelity of both the recovered saliency and the watermarked background gradually decreases. We observe that fidelity remains at a high level when the facial region percentage is $60\%$ or less. Beyond this, the decline becomes more obvious, indicating that a larger payload (i.e., a bigger face) strains the limited capacity of the background, consequently leading to fidelity reduction.

\noindent \textbf{Ownership Verification.} As described in Section \ref{sec:recovery_localization_verification}, we adopt a NCC threshold of $0.95$ for ownership verification. Across all test settings, RecoverMark achieves a Success Rate of $99.9\%$ in ownership verification, demonstrating its effectiveness as a copyright verification mechanism.

\begin{table*}[!t]
\centering
\caption{Quantitative comparison on the ID dataset (CelebA \cite{Dataset:CelebA}) under Structpix2pix \cite{brooks2022instructpix2pix} manipulation.}
\label{tab:CelebA_Structpix2pix}

\resizebox{\textwidth}{!}{%
    \begin{tabular}{
        @{}c|c|
        c c | 
        c c | 
        c c | 
        c c | 
        c c | 
        c c     
        @{}
    }
        \toprule
        \toprule
        \multicolumn{2}{c|}{\multirow{3}{*}{\textbf{Method}}} & \multicolumn{12}{c|}{\textbf{Localization}}  \\
        \cmidrule(l){3-14}
        \multicolumn{2}{c|}{} & \multicolumn{2}{c|}{\textbf{Regeneration}} & \multicolumn{2}{c|}{\textbf{Patch Remove}} & \multicolumn{2}{c|}{\textbf{Noise}} & \multicolumn{2}{c|}{\textbf{JPEG}} & \multicolumn{2}{c|}{\textbf{LP Filter}} & \multicolumn{2}{c}{\textbf{Lattice \cite{Liu2023Erase_Removal_Attack}}} \\
        \cmidrule(l){3-14}
        \multicolumn{2}{c|}{} & {\textbf{F1}} & {\textbf{AUC}} & {\textbf{F1}} & {\textbf{AUC}} & {\textbf{F1}} & {\textbf{AUC}} & {\textbf{F1}} & {\textbf{AUC}} & {\textbf{F1}} & {\textbf{AUC}} & {\textbf{F1}} & {\textbf{AUC}} \\
        \midrule
        \multirow{2}{*}{\rotatebox{90}{\textbf{Passive}}}
        & MVSS-Net \cite{PassiveMVSSNet}   
        & \databarRatio[red!25]{0.041} & \databarRatio[blue!25]{0.723} 
        & \databarRatio[red!25]{0.025} & \databarRatio[blue!25]{0.685} 
        & \databarRatio[red!25]{0.062} & \databarRatio[blue!25]{0.711} 
        & \databarRatio[red!25]{0.157}  & \databarRatio[blue!25]{0.776} 
        & \databarRatio[red!25]{0.184}  & \databarRatio[blue!25]{0.755} 
        & \databarRatio[red!25]{0.034}  & \databarRatio[blue!25]{0.719} \\
        \cmidrule{2-14}
        & HiFiNet \cite{PassiveHiFiNet}   
        & \databarRatio[red!25]{0.048} & \databarRatio[blue!25]{0.522} 
        & \databarRatio[red!25]{0.020} & \databarRatio[blue!25]{0.531} 
        & \databarRatio[red!25]{0.022} & \databarRatio[blue!25]{0.553} 
        & \databarRatio[red!25]{0.080}  & \databarRatio[blue!25]{0.608} 
        & \databarRatio[red!25]{0.067}  & \databarRatio[blue!25]{0.552} 
        & \databarRatio[red!25]{0.051}  & \databarRatio[blue!25]{0.556} \\
        \midrule
        \multirow{5}{*}{\rotatebox{90}{\textbf{Proactive}}}
        & Imuge+ \cite{ProactiveImmunePlus}    
        & \databarRatio[red!25]{0.091} & \databarRatio[blue!25]{0.587} 
        & \databarRatio[red!25]{0.217} & \databarRatio[blue!25]{0.835} 
        & \databarRatio[red!25]{0.016} & \databarRatio[blue!25]{0.639} 
        & \databarRatio[red!25]{0.419}  & \databarRatio[blue!25]{0.956} 
        & \databarRatio[red!25]{0.087}  & \databarRatio[blue!25]{0.530} 
        & \databarRatio[red!25]{0.010}  & \databarRatio[blue!25]{0.585} \\
        \cmidrule{2-14}
        & EditGuard \cite{BaselineEditGuard} 
        & \databarRatio[red!25]{0.090} & \databarRatio[blue!25]{0.610} 
        & \databarRatio[red!25]{0.243} & \databarRatio[blue!25]{0.888} 
        & \databarRatio[red!25]{0.528} & \databarRatio[blue!25]{0.932} 
        & \databarRatio[red!25]{0.552}  & \databarRatio[blue!25]{0.954} 
        & \databarRatio[red!25]{0.090}  & \databarRatio[blue!25]{0.658} 
        & \databarRatio[red!25]{0.438}  & \databarRatio[blue!25]{0.930} \\
        \cmidrule{2-14}
        & OmniGuard \cite{BaselineOmniGuard} 
        & \databarRatio[red!25]{0.105} & \databarRatio[blue!25]{0.655} 
        & \databarRatio[red!25]{0.535} & \databarRatio[blue!25]{0.960} 
        & \databarRatio[red!25]{0.127} & \databarRatio[blue!25]{0.659} 
        & \databarRatio[red!25]{0.315}  & \databarRatio[blue!25]{0.890} 
        & \databarRatio[red!25]{0.146}  & \databarRatio[blue!25]{0.743} 
        & \databarRatio[red!25]{0.113}  & \databarRatio[blue!25]{0.689} \\
        \cmidrule{2-14}
        & RecoverMark (Ours) 
        & \databarRatio[red!25]{0.855} & \databarRatio[blue!25]{0.993} 
        & \databarRatio[red!25]{0.564} & \databarRatio[blue!25]{0.974} 
        & \databarRatio[red!25]{0.876} & \databarRatio[blue!25]{0.992} 
        & \databarRatio[red!25]{0.867}  & \databarRatio[blue!25]{0.993} 
        & \databarRatio[red!25]{0.830}  & \databarRatio[blue!25]{0.989} 
        & \databarRatio[red!25]{0.842}  & \databarRatio[blue!25]{0.991} \\
        \midrule[1.5pt]
        \multicolumn{2}{c|}{\multirow{3}{*}{\textbf{Method}}} & \multicolumn{12}{c}{\textbf{Recovery}} \\
        \cmidrule(l){3-14}
        \multicolumn{2}{c|}{} & \multicolumn{2}{c|}{\textbf{Regeneration}} & \multicolumn{2}{c|}{\textbf{Patch Remove}} & \multicolumn{2}{c|}{\textbf{Noise}} & \multicolumn{2}{c|}{\textbf{JPEG}} & \multicolumn{2}{c|}{\textbf{LP Filter}} & \multicolumn{2}{c}{\textbf{Lattice \cite{Liu2023Erase_Removal_Attack}}} \\
        \cmidrule(l){3-14}
        \multicolumn{2}{c|}{} & \textbf{PSNR} & \textbf{MS-SSIM} & \textbf{PSNR} & \textbf{MS-SSIM} & \textbf{PSNR} & \textbf{MS-SSIM} & \textbf{PSNR} & \textbf{MS-SSIM} & \textbf{PSNR} & \textbf{MS-SSIM} & \textbf{PSNR} & \textbf{MS-SSIM} \\
        \midrule
        \multicolumn{2}{c|}{Imuge+ \cite{ProactiveImmunePlus}}
        & \databarPSNR[red!25]{7.252} & \databarRatio[blue!25]{0.339}
        & \databarPSNR[red!25]{14.445} & \databarRatio[blue!25]{0.673}
        & \databarPSNR[red!25]{10.432} & \databarRatio[blue!25]{0.563}
        & \databarPSNR[red!25]{10.778} & \databarRatio[blue!25]{0.629}
        & \databarPSNR[red!25]{10.972} & \databarRatio[blue!25]{0.603}
        & \databarPSNR[red!25]{9.089} & \databarRatio[blue!25]{0.424} \\
        \cmidrule{1-14}
        \multicolumn{2}{c|}{RecoverMark (Ours)}
        & \databarPSNR[red!25]{22.154} & \databarRatio[blue!25]{0.607}
        & \databarPSNR[red!25]{16.624} & \databarRatio[blue!25]{0.467}
        & \databarPSNR[red!25]{23.276} & \databarRatio[blue!25]{0.657}
        & \databarPSNR[red!25]{23.314} & \databarRatio[blue!25]{0.680}
        & \databarPSNR[red!25]{21.744} & \databarRatio[blue!25]{0.660}
        & \databarPSNR[red!25]{23.230} & \databarRatio[blue!25]{0.655} \\
        \bottomrule
        \bottomrule
    \end{tabular}%
}
\end{table*}

\begin{table*}[!t]
\centering
\caption{Quantitative comparison on the ID dataset (CelebA \cite{Dataset:CelebA}) under Stable Diffusion \cite{StableDiffusion} Inpaint manipulation.}
\label{tab:CelebA_sd_inpaint}

\resizebox{\textwidth}{!}{%
    \begin{tabular}{
        @{}c|c|
        c c | 
        c c | 
        c c | 
        c c | 
        c c | 
        c c     
        @{}
    }
        \toprule
        \toprule
        \multicolumn{2}{c|}{\multirow{3}{*}{\textbf{Method}}} & \multicolumn{12}{c|}{\textbf{Localization}}  \\
        \cmidrule(l){3-14}
        \multicolumn{2}{c|}{} & \multicolumn{2}{c|}{\textbf{Regeneration}} & \multicolumn{2}{c|}{\textbf{Patch Remove}} & \multicolumn{2}{c|}{\textbf{Noise}} & \multicolumn{2}{c|}{\textbf{JPEG}} & \multicolumn{2}{c|}{\textbf{LP Filter}} & \multicolumn{2}{c}{\textbf{Lattice \cite{Liu2023Erase_Removal_Attack}}} \\
        \cmidrule(l){3-14}
        \multicolumn{2}{c|}{} & {\textbf{F1}} & {\textbf{AUC}} & {\textbf{F1}} & {\textbf{AUC}} & {\textbf{F1}} & {\textbf{AUC}} & {\textbf{F1}} & {\textbf{AUC}} & {\textbf{F1}} & {\textbf{AUC}} & {\textbf{F1}} & {\textbf{AUC}} \\
        \midrule
        \multirow{2}{*}{\rotatebox{90}{\textbf{Passive}}}
        & MVSS-Net \cite{PassiveMVSSNet} 
        & \databarRatio[red!25]{0.257} & \databarRatio[blue!25]{0.810} 
        & \databarRatio[red!25]{0.245} & \databarRatio[blue!25]{0.786} 
        & \databarRatio[red!25]{0.139} & \databarRatio[blue!25]{0.750} 
        & \databarRatio[red!25]{0.213}  & \databarRatio[blue!25]{0.776} 
        & \databarRatio[red!25]{0.280}  & \databarRatio[blue!25]{0.726} 
        & \databarRatio[red!25]{0.050}  & \databarRatio[blue!25]{0.700} \\
        \cmidrule{2-14}
        & HiFiNet \cite{PassiveHiFiNet}   
        & \databarRatio[red!25]{0.058} & \databarRatio[blue!25]{0.635} 
        & \databarRatio[red!25]{0.041} & \databarRatio[blue!25]{0.621} 
        & \databarRatio[red!25]{0.002} & \databarRatio[blue!25]{0.544} 
        & \databarRatio[red!25]{0.050}  & \databarRatio[blue!25]{0.607} 
        & \databarRatio[red!25]{0.091}  & \databarRatio[blue!25]{0.695} 
        & \databarRatio[red!25]{0.031}  & \databarRatio[blue!25]{0.625} \\
        \midrule
        \multirow{5}{*}{\rotatebox{90}{\textbf{Proactive}}}
        & Imuge+ \cite{ProactiveImmunePlus}    
        & \databarRatio[red!25]{0.195} & \databarRatio[blue!25]{0.782} 
        & \databarRatio[red!25]{0.442} & \databarRatio[blue!25]{0.915} 
        & \databarRatio[red!25]{0.012} & \databarRatio[blue!25]{0.655} 
        & \databarRatio[red!25]{0.983}  & \databarRatio[blue!25]{0.862} 
        & \databarRatio[red!25]{0.186}  & \databarRatio[blue!25]{0.628} 
        & \databarRatio[red!25]{0.034}  & \databarRatio[blue!25]{0.546} \\
        \cmidrule{2-14}
        & EditGuard \cite{BaselineEditGuard} 
        & \databarRatio[red!25]{0.176} & \databarRatio[blue!25]{0.580} 
        & \databarRatio[red!25]{0.434} & \databarRatio[blue!25]{0.970} 
        & \databarRatio[red!25]{0.792} & \databarRatio[blue!25]{0.990} 
        & \databarRatio[red!25]{0.835}  & \databarRatio[blue!25]{0.995} 
        & \databarRatio[red!25]{0.176}  & \databarRatio[blue!25]{0.580} 
        & \databarRatio[red!25]{0.704}  & \databarRatio[blue!25]{0.989} \\
        \cmidrule{2-14}
        & OmniGuard \cite{BaselineOmniGuard} 
        & \databarRatio[red!25]{0.064} & \databarRatio[blue!25]{0.536} 
        & \databarRatio[red!25]{0.911} & \databarRatio[blue!25]{0.999} 
        & \databarRatio[red!25]{0.231} & \databarRatio[blue!25]{0.693} 
        & \databarRatio[red!25]{0.547}  & \databarRatio[blue!25]{0.920} 
        & \databarRatio[red!25]{0.273}  & \databarRatio[blue!25]{0.753} 
        & \databarRatio[red!25]{0.275}  & \databarRatio[blue!25]{0.757} \\
        \cmidrule{2-14}
        & RecoverMark (Ours) 
        & \databarRatio[red!25]{0.721} & \databarRatio[blue!25]{0.992} 
        & \databarRatio[red!25]{0.569} & \databarRatio[blue!25]{0.977} 
        & \databarRatio[red!25]{0.759} & \databarRatio[blue!25]{0.994} 
        & \databarRatio[red!25]{0.771}  & \databarRatio[blue!25]{0.994} 
        & \databarRatio[red!25]{0.699}  & \databarRatio[blue!25]{0.989} 
        & \databarRatio[red!25]{0.756}  & \databarRatio[blue!25]{0.993} \\
        \midrule[1.5pt]
        \multicolumn{2}{c|}{\multirow{3}{*}{\textbf{Method}}} & \multicolumn{12}{c}{\textbf{Recovery}} \\
        \cmidrule(l){3-14}
        \multicolumn{2}{c|}{} & \multicolumn{2}{c|}{\textbf{Regeneration}} & \multicolumn{2}{c|}{\textbf{Patch Remove}} & \multicolumn{2}{c|}{\textbf{Noise}} & \multicolumn{2}{c|}{\textbf{JPEG}} & \multicolumn{2}{c|}{\textbf{LP Filter}} & \multicolumn{2}{c}{\textbf{Lattice \cite{Liu2023Erase_Removal_Attack}}} \\
        \cmidrule(l){3-14}
        \multicolumn{2}{c|}{} & \textbf{PSNR} & \textbf{MS-SSIM} & \textbf{PSNR} & \textbf{MS-SSIM} & \textbf{PSNR} & \textbf{MS-SSIM} & \textbf{PSNR} & \textbf{MS-SSIM} & \textbf{PSNR} & \textbf{MS-SSIM} & \textbf{PSNR} & \textbf{MS-SSIM} \\
        \midrule
        \multicolumn{2}{c|}{Imuge+ \cite{ProactiveImmunePlus}}
        & \databarPSNR[red!25]{11.997} & \databarRatio[blue!25]{0.318}
        & \databarPSNR[red!25]{22.673} & \databarRatio[blue!25]{0.702}
        & \databarPSNR[red!25]{18.296} & \databarRatio[blue!25]{0.544}
        & \databarPSNR[red!25]{21.544} & \databarRatio[blue!25]{0.671}
        & \databarPSNR[red!25]{13.865} & \databarRatio[blue!25]{0.397}
        & \databarPSNR[red!25]{17.643} & \databarRatio[blue!25]{0.487} \\
        \cmidrule{1-14}
        \multicolumn{2}{c|}{RecoverMark (Ours)}
        & \databarPSNR[red!25]{22.553} & \databarRatio[blue!25]{0.610}
        & \databarPSNR[red!25]{17.236} & \databarRatio[blue!25]{0.491}
        & \databarPSNR[red!25]{23.421} & \databarRatio[blue!25]{0.659}
        & \databarPSNR[red!25]{23.745} & \databarRatio[blue!25]{0.681}
        & \databarPSNR[red!25]{21.113} & \databarRatio[blue!25]{0.621}
        & \databarPSNR[red!25]{23.304} & \databarRatio[blue!25]{0.656} \\
        \bottomrule
        \bottomrule
    \end{tabular}%
}
\end{table*}

\begin{table*}[!t]
\centering
\caption{Quantitative comparison on the OOD dataset (FFHQ \cite{Dataset:FFHQ}) under Instructpix2pix \cite{brooks2022instructpix2pix} manipulation.}
\label{tab:FFHQ_Structpix2pix}

\resizebox{\textwidth}{!}{%
    \begin{tabular}{
        @{}c|c|
        c c | 
        c c | 
        c c | 
        c c | 
        c c | 
        c c     
        @{}
    }
        \toprule
        \toprule
        \multicolumn{2}{c|}{\multirow{3}{*}{\textbf{Method}}} & \multicolumn{12}{c|}{\textbf{Localization}}  \\
        \cmidrule(l){3-14}
        \multicolumn{2}{c|}{} & \multicolumn{2}{c|}{\textbf{Regeneration}} & \multicolumn{2}{c|}{\textbf{Patch Remove}} & \multicolumn{2}{c|}{\textbf{Noise}} & \multicolumn{2}{c|}{\textbf{JPEG}} & \multicolumn{2}{c|}{\textbf{LP Filter}} & \multicolumn{2}{c}{\textbf{Lattice \cite{Liu2023Erase_Removal_Attack}}} \\
        \cmidrule(l){3-14}
        \multicolumn{2}{c|}{} & {\textbf{F1}} & {\textbf{AUC}} & {\textbf{F1}} & {\textbf{AUC}} & {\textbf{F1}} & {\textbf{AUC}} & {\textbf{F1}} & {\textbf{AUC}} & {\textbf{F1}} & {\textbf{AUC}} & {\textbf{F1}} & {\textbf{AUC}} \\
        \midrule
        \multirow{2}{*}{\rotatebox{90}{\textbf{Passive}}}
        & MVSS-Net \cite{PassiveMVSSNet}   
        & \databarRatio[red!25]{0.076} & \databarRatio[blue!25]{0.686} 
        & \databarRatio[red!25]{0.042} & \databarRatio[blue!25]{0.668} 
        & \databarRatio[red!25]{0.049} & \databarRatio[blue!25]{0.678} 
        & \databarRatio[red!25]{0.171}  & \databarRatio[blue!25]{0.776} 
        & \databarRatio[red!25]{0.241}  & \databarRatio[blue!25]{0.803} 
        & \databarRatio[red!25]{0.028}  & \databarRatio[blue!25]{0.652} \\
        \cmidrule{2-14}
        & HiFiNet \cite{PassiveHiFiNet}   
        & \databarRatio[red!25]{0.059} & \databarRatio[blue!25]{0.511} 
        & \databarRatio[red!25]{0.003} & \databarRatio[blue!25]{0.512} 
        & \databarRatio[red!25]{0.025} & \databarRatio[blue!25]{0.526} 
        & \databarRatio[red!25]{0.063}  & \databarRatio[blue!25]{0.537} 
        & \databarRatio[red!25]{0.024}  & \databarRatio[blue!25]{0.503} 
        & \databarRatio[red!25]{0.050}  & \databarRatio[blue!25]{0.561} \\
        \midrule
        \multirow{5}{*}{\rotatebox{90}{\textbf{Proactive}}}
        & Imuge+ \cite{ProactiveImmunePlus}    
        & \databarRatio[red!25]{0.081} & \databarRatio[blue!25]{0.556} 
        & \databarRatio[red!25]{0.269} & \databarRatio[blue!25]{0.867} 
        & \databarRatio[red!25]{0.045} & \databarRatio[blue!25]{0.593} 
        & \databarRatio[red!25]{0.463}  & \databarRatio[blue!25]{0.962} 
        & \databarRatio[red!25]{0.142}  & \databarRatio[blue!25]{0.763} 
        & \databarRatio[red!25]{0.030}  & \databarRatio[blue!25]{0.546} \\
        \cmidrule{2-14}
        & EditGuard \cite{BaselineEditGuard} 
        & \databarRatio[red!25]{0.108} & \databarRatio[blue!25]{0.612} 
        & \databarRatio[red!25]{0.286} & \databarRatio[blue!25]{0.897} 
        & \databarRatio[red!25]{0.572} & \databarRatio[blue!25]{0.940} 
        & \databarRatio[red!25]{0.601}  & \databarRatio[blue!25]{0.957} 
        & \databarRatio[red!25]{0.115}  & \databarRatio[blue!25]{0.561} 
        & \databarRatio[red!25]{0.483}  & \databarRatio[blue!25]{0.932} \\
        \cmidrule{2-14}
        & OmniGuard \cite{BaselineOmniGuard} 
        & \databarRatio[red!25]{0.124} & \databarRatio[blue!25]{0.664} 
        & \databarRatio[red!25]{0.599} & \databarRatio[blue!25]{0.975} 
        & \databarRatio[red!25]{0.147} & \databarRatio[blue!25]{0.684} 
        & \databarRatio[red!25]{0.333}  & \databarRatio[blue!25]{0.882} 
        & \databarRatio[red!25]{0.134}  & \databarRatio[blue!25]{0.660} 
        & \databarRatio[red!25]{0.127}  & \databarRatio[blue!25]{0.676} \\
        \cmidrule{2-14}
        & RecoverMark (Ours) 
        & \databarRatio[red!25]{0.884} & \databarRatio[blue!25]{0.997} 
        & \databarRatio[red!25]{0.585} & \databarRatio[blue!25]{0.978} 
        & \databarRatio[red!25]{0.904} & \databarRatio[blue!25]{0.998} 
        & \databarRatio[red!25]{0.893}  & \databarRatio[blue!25]{0.996} 
        & \databarRatio[red!25]{0.853}  & \databarRatio[blue!25]{0.995} 
        & \databarRatio[red!25]{0.882}  & \databarRatio[blue!25]{0.997} \\
        \midrule[1.5pt]
        \multicolumn{2}{c|}{\multirow{3}{*}{\textbf{Method}}} & \multicolumn{12}{c}{\textbf{Recovery}} \\
        \cmidrule(l){3-14}
        \multicolumn{2}{c|}{} & \multicolumn{2}{c|}{\textbf{Regeneration}} & \multicolumn{2}{c|}{\textbf{Patch Remove}} & \multicolumn{2}{c|}{\textbf{Noise}} & \multicolumn{2}{c|}{\textbf{JPEG}} & \multicolumn{2}{c|}{\textbf{LP Filter}} & \multicolumn{2}{c}{\textbf{Lattice \cite{Liu2023Erase_Removal_Attack}}} \\
        \cmidrule(l){3-14}
        \multicolumn{2}{c|}{} & \textbf{PSNR} & \textbf{MS-SSIM} & \textbf{PSNR} & \textbf{MS-SSIM} & \textbf{PSNR} & \textbf{MS-SSIM} & \textbf{PSNR} & \textbf{MS-SSIM} & \textbf{PSNR} & \textbf{MS-SSIM} & \textbf{PSNR} & \textbf{MS-SSIM} \\
        \midrule
        \multicolumn{2}{c|}{Imuge+ \cite{ProactiveImmunePlus} }
        & \databarPSNR[red!25]{9.531} & \databarRatio[blue!25]{0.327}
        & \databarPSNR[red!25]{14.075} & \databarRatio[blue!25]{0.590}
        & \databarPSNR[red!25]{14.040} & \databarRatio[blue!25]{0.541}
        & \databarPSNR[red!25]{15.332} & \databarRatio[blue!25]{0.608}
        & \databarPSNR[red!25]{12.081} & \databarRatio[blue!25]{0.517}
        & \databarPSNR[red!25]{13.427} & \databarRatio[blue!25]{0.421} \\
        \cmidrule{1-14}
        \multicolumn{2}{c|}{RecoverMark (Ours)}
        & \databarPSNR[red!25]{22.052} & \databarRatio[blue!25]{0.593}
        & \databarPSNR[red!25]{16.951} & \databarRatio[blue!25]{0.476}
        & \databarPSNR[red!25]{23.215} & \databarRatio[blue!25]{0.642}
        & \databarPSNR[red!25]{23.124} & \databarRatio[blue!25]{0.657}
        & \databarPSNR[red!25]{20.618} & \databarRatio[blue!25]{0.607}
        & \databarPSNR[red!25]{22.993} & \databarRatio[blue!25]{0.637} \\
        \bottomrule
        \bottomrule
    \end{tabular}%
}
\end{table*}

\begin{table*}[!t]
\centering
\caption{Quantitative comparison on the OOD dataset (FFHQ \cite{Dataset:FFHQ}) under Stable Diffusion \cite{StableDiffusion} Inpaint manipulation.}
\label{tab:FFHQ_sd_inpaint}

\resizebox{\textwidth}{!}{%
    \begin{tabular}{
        @{}c|c|
        c c | 
        c c | 
        c c | 
        c c | 
        c c | 
        c c     
        @{}
    }
        \toprule
        \toprule
        \multicolumn{2}{c|}{\multirow{3}{*}{\textbf{Method}}} & \multicolumn{12}{c|}{\textbf{Localization}}  \\
        \cmidrule(l){3-14}
        \multicolumn{2}{c|}{} & \multicolumn{2}{c|}{\textbf{Regeneration}} & \multicolumn{2}{c|}{\textbf{Patch Remove}} & \multicolumn{2}{c|}{\textbf{Noise}} & \multicolumn{2}{c|}{\textbf{JPEG}} & \multicolumn{2}{c|}{\textbf{LP Filter}} & \multicolumn{2}{c}{\textbf{Lattice \cite{Liu2023Erase_Removal_Attack}}} \\
        \cmidrule(l){3-14}
        \multicolumn{2}{c|}{} & {\textbf{F1}} & {\textbf{AUC}} & {\textbf{F1}} & {\textbf{AUC}} & {\textbf{F1}} & {\textbf{AUC}} & {\textbf{F1}} & {\textbf{AUC}} & {\textbf{F1}} & {\textbf{AUC}} & {\textbf{F1}} & {\textbf{AUC}} \\
        \midrule
        \multirow{2}{*}{\rotatebox{90}{\textbf{Passive}}}
        & MVSS-Net \cite{PassiveMVSSNet}   
        & \databarRatio[red!25]{0.215} & \databarRatio[blue!25]{0.805} 
        & \databarRatio[red!25]{0.155} & \databarRatio[blue!25]{0.813} 
        & \databarRatio[red!25]{0.071} & \databarRatio[blue!25]{0.773} 
        & \databarRatio[red!25]{0.189}  & \databarRatio[blue!25]{0.789} 
        & \databarRatio[red!25]{0.348}  & \databarRatio[blue!25]{0.763} 
        & \databarRatio[red!25]{0.012}  & \databarRatio[blue!25]{0.713} \\
        \cmidrule{2-14}
        & HiFiNet \cite{PassiveHiFiNet}   
        & \databarRatio[red!25]{0.071} & \databarRatio[blue!25]{0.628} 
        & \databarRatio[red!25]{0.041} & \databarRatio[blue!25]{0.621} 
        & \databarRatio[red!25]{0.002} & \databarRatio[blue!25]{0.605} 
        & \databarRatio[red!25]{0.068}  & \databarRatio[blue!25]{0.648} 
        & \databarRatio[red!25]{0.128}  & \databarRatio[blue!25]{0.706} 
        & \databarRatio[red!25]{0.031}  & \databarRatio[blue!25]{0.659} \\
        \midrule
        \multirow{5}{*}{\rotatebox{90}{\textbf{Proactive}}}
        & Imuge+ \cite{ProactiveImmunePlus}    
        & \databarRatio[red!25]{0.219} & \databarRatio[blue!25]{0.774} 
        & \databarRatio[red!25]{0.470} & \databarRatio[blue!25]{0.924} 
        & \databarRatio[red!25]{0.012} & \databarRatio[blue!25]{0.589} 
        & \databarRatio[red!25]{0.672}  & \databarRatio[blue!25]{0.987} 
        & \databarRatio[red!25]{0.207}  & \databarRatio[blue!25]{0.661} 
        & \databarRatio[red!25]{0.008}  & \databarRatio[blue!25]{0.537} \\
        \cmidrule{2-14}
        & EditGuard \cite{BaselineEditGuard} 
        & \databarRatio[red!25]{0.196} & \databarRatio[blue!25]{0.583} 
        & \databarRatio[red!25]{0.458} & \databarRatio[blue!25]{0.968} 
        & \databarRatio[red!25]{0.800} & \databarRatio[blue!25]{0.989} 
        & \databarRatio[red!25]{0.839}  & \databarRatio[blue!25]{0.996} 
        & \databarRatio[red!25]{0.196}  & \databarRatio[blue!25]{0.758} 
        & \databarRatio[red!25]{0.711}  & \databarRatio[blue!25]{0.987} \\
        \cmidrule{2-14}
        & OmniGuard \cite{BaselineOmniGuard} 
        & \databarRatio[red!25]{0.152} & \databarRatio[blue!25]{0.611} 
        & \databarRatio[red!25]{0.895} & \databarRatio[blue!25]{0.999} 
        & \databarRatio[red!25]{0.271} & \databarRatio[blue!25]{0.735} 
        & \databarRatio[red!25]{0.552}  & \databarRatio[blue!25]{0.913} 
        & \databarRatio[red!25]{0.254}  & \databarRatio[blue!25]{0.696} 
        & \databarRatio[red!25]{0.235}  & \databarRatio[blue!25]{0.7778} \\
        \cmidrule{2-14}
        & RecoverMark (Ours) 
        & \databarRatio[red!25]{0.732} & \databarRatio[blue!25]{0.993} 
        & \databarRatio[red!25]{0.603} & \databarRatio[blue!25]{0.976} 
        & \databarRatio[red!25]{0.753} & \databarRatio[blue!25]{0.993} 
        & \databarRatio[red!25]{0.759}  & \databarRatio[blue!25]{0.994} 
        & \databarRatio[red!25]{0.716}  & \databarRatio[blue!25]{0.987} 
        & \databarRatio[red!25]{0.754}  & \databarRatio[blue!25]{0.992} \\
        \midrule[1.5pt]
        \multicolumn{2}{c|}{\multirow{3}{*}{\textbf{Method}}} & \multicolumn{12}{c}{\textbf{Recovery}} \\
        \cmidrule(l){3-14}
        \multicolumn{2}{c|}{} & \multicolumn{2}{c|}{\textbf{Regeneration}} & \multicolumn{2}{c|}{\textbf{Patch Remove}} & \multicolumn{2}{c|}{\textbf{Noise}} & \multicolumn{2}{c|}{\textbf{JPEG}} & \multicolumn{2}{c|}{\textbf{LP Filter}} & \multicolumn{2}{c}{\textbf{Lattice \cite{Liu2023Erase_Removal_Attack}}} \\
        \cmidrule(l){3-14}
        \multicolumn{2}{c|}{} & \textbf{PSNR} & \textbf{MS-SSIM} & \textbf{PSNR} & \textbf{MS-SSIM} & \textbf{PSNR} & \textbf{MS-SSIM} & \textbf{PSNR} & \textbf{MS-SSIM} & \textbf{PSNR} & \textbf{MS-SSIM} & \textbf{PSNR} & \textbf{MS-SSIM} \\
        \midrule
        \multicolumn{2}{c|}{Imuge+ \cite{ProactiveImmunePlus}}
        & \databarPSNR[red!25]{11.975} & \databarRatio[blue!25]{0.314}
        & \databarPSNR[red!25]{22.51} & \databarRatio[blue!25]{0.691}
        & \databarPSNR[red!25]{17.993} & \databarRatio[blue!25]{0.531}
        & \databarPSNR[red!25]{21.500} & \databarRatio[blue!25]{0.663}
        & \databarPSNR[red!25]{13.964} & \databarRatio[blue!25]{0.399}
        & \databarPSNR[red!25]{17.464} & \databarRatio[blue!25]{0.474} \\
        \cmidrule{1-14}
        \multicolumn{2}{c|}{RecoverMark (Ours)}
        & \databarPSNR[red!25]{22.103} & \databarRatio[blue!25]{0.591}
        & \databarPSNR[red!25]{17.225} & \databarRatio[blue!25]{0.487}
        & \databarPSNR[red!25]{23.221} & \databarRatio[blue!25]{0.641}
        & \databarPSNR[red!25]{23.176} & \databarRatio[blue!25]{0.657}
        & \databarPSNR[red!25]{20.623} & \databarRatio[blue!25]{0.607}
        & \databarPSNR[red!25]{23.031} & \databarRatio[blue!25]{0.638} \\
        \bottomrule
        \bottomrule
    \end{tabular}%
}
\end{table*}

\begin{figure}[!t]
    \centering
    \includegraphics[width=0.95\linewidth]{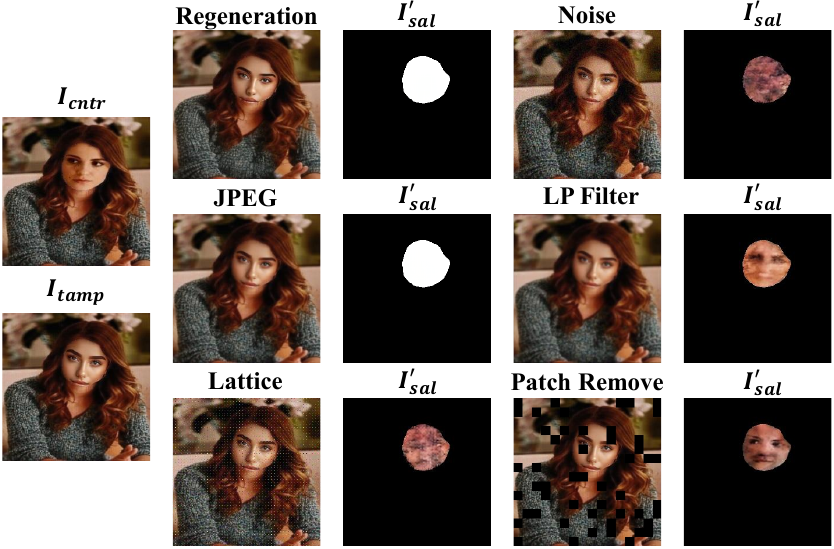}
    \caption{Ablation study on the contribution of the distortion layer to RecoverMark.}
    \label{fig:without_distortion}
\end{figure}

\begin{figure}[!t]
    \centering
    \includegraphics[width=0.9\linewidth]{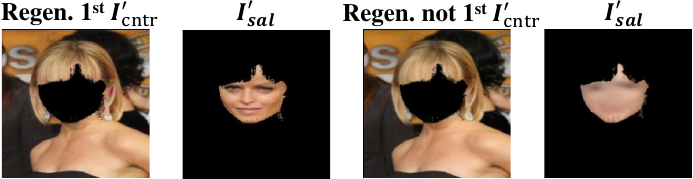}
    \caption{Ablation study on the impact of the distortion order.}
    \label{fig:distortion_order}
\end{figure}

\subsection{Ablation Study}
\label{sec:ablation_study}
We conduct two ablation studies to analyze the impact of the distortion layer and the distortion order. On the one hand, as illustrated in Figure~\ref{fig:without_distortion}, RecoverMark trained with only Stage 1 (no distortion layer applied) fails to clearly recover the original face under removal attacks. On the other hand, as demonstrated in Figure~\ref{fig:distortion_order}, we find that the distortion order is crucial and observe that introducing the regeneration attack as the first distortion in the progressive training pipeline yields superior results. However, if it is placed later in the sequence, RecoverMark fails to develop effective robustness against these attacks. Our conjecture is that employing the most powerful regeneration attack first sets a high standard for robustness early on, guiding the model towards a globally robust embedding rather than allowing it to converge on a ``lazy" local minimum effective only against simpler distortions.

\section{Further Discussion}
While RecoverMark demonstrates significant robustness against watermark removal attacks and achieves image manipulation localization, recovery, and ownership verification at the same time, in this paper, we limit the scope only to facial region. However, it is easy to generalize the idea to protect other object categories by segmenting a different salient object as the watermark and using its corresponding background as the container. Regarding limitation, the method is limited by an inherent capacity trade-off. As discussed in Section~\ref{sec:quantitative_evaluation}, in extreme cases, such as close-up portraits where the facial region is large and the background area is minimal, the limited background capacity may be insufficient to store a high-fidelity watermark, reducing both embedded background and quality of the recovered image. Therefore, higher-capacity robust watermarking techniques also merit further investigation.

\section{Conclusion}
In this paper, we have introduced RecoverMark, a robust framework that addresses the overlooked vulnerability of existing methods when subjected to watermark removal attacks prior to manipulation. Our method leverages the practical constraint of background consistency, treating the face content itself as a watermark and embedding it into the background to achieve image manipulation localization, recovery, and ownership verification simultaneously. In addition, we have proposed a two-stage, progressive training pipeline that further enhances RecoverMark's robustness. Extensive experiments show that RecoverMark successfully achieves manipulation localization, recovery, and ownership verification, demonstrating strong generalizability to both seen and unseen attacks, as well as ID and OOD data. We believe that RecoverMark offers significant insights and will inspire future work extending this paradigm to generic images with non-facial saliency.

\section*{Acknowledgment}
The research work described in this paper was conducted in the JC STEM Lab of Smart City funded by The Hong Kong Jockey Club Charities Trust under Contract 2023-0108. This work was also supported in part by the Hong Kong SAR Government under the Global STEM Professorship and Research Talent Hub, by Singapore Ministry of Education (MOE) under the Academic Research Fund (AcRF) Tier 1 Grant R-MA123-R205-0008, and by the National Research Foundation Singapore under the AI Singapore Programme (AISG Award No: AISG3-RPGV-2025-019). We sincerely appreciate the support provided.

{
    \small
    \bibliographystyle{ieeenat_fullname}
    \bibliography{main}
}
\end{document}